\documentclass{article} 
\usepackage[main,preprint]{neurips_2025}
\makeatletter
\renewcommand{\@noticestring}{\leavevmode\phantom{\@trackname}}
\makeatother
\usepackage[T1]{fontenc}
\usepackage{microtype}
\setcitestyle{authoryear,round,citesep={;},aysep={,},yysep={;}}

\usepackage[table]{xcolor}
\usepackage{booktabs}
\usepackage{multirow}
\usepackage{graphicx}
\usepackage{amsmath}
\usepackage{amssymb}
\usepackage[hidelinks]{hyperref}
\usepackage{url}

\usepackage{adjustbox}

\usepackage[skins,breakable]{tcolorbox}
\usepackage{tabularx}
\usepackage{array}

\let\cite\citep

\definecolor{LightCyan}{rgb}{0.88,1,1}
\definecolor{LightRed}{rgb}{1,0.88,1}
\definecolor{LightYellow}{rgb}{1,1,0.88}
\definecolor{LightGray}{gray}{0.8}

\usepackage{cleveref}
\crefformat{section}{\S#2#1#3} 
\crefformat{subsection}{\S#2#1#3}

\tcbset{
    findingbox/.style={
    width=\linewidth,
    top=8pt,
    bottom=4pt,
    colback=blue!6!white,
    colframe=black,
    colbacktitle=black,
    enhanced,
    center,
    attach boxed title to top left={yshift=-0.1in,xshift=0.15in},
    boxed title style={boxrule=0pt,colframe=white,},
  }
}
\newtcolorbox[auto counter]{Findingbox}[2][]{%
  findingbox,
  title={Finding~\thetcbcounter. #2},
  label type=finding,
  #1%
}

\definecolor{blocktitlebg}{RGB}{134,145,200} 
\definecolor{blockbodybg}{RGB}{243,244,250} 
\definecolor{blockline}{RGB}{110,122,185}  

\newtcolorbox{defblock}[2][]{
  enhanced,
  colback=blockbodybg,
  colbacktitle=blocktitlebg,
  colframe=blockline,
  coltitle=white,
  fonttitle=\bfseries,
  boxrule=0.3pt,
  arc=0.5mm,
  left=3mm, right=2mm, top=1mm, bottom=1mm,
  title=#2, #1
}

\usepackage{fancyvrb}

\definecolor{PromptBlue}{HTML}{49649B}
\definecolor{PromptBg}{HTML}{F7F9FC}
\definecolor{PromptLine}{HTML}{D9E1ED}

\newtcolorbox{promptcard}[1]{%
  enhanced, breakable,
  colback=PromptBg,
  colframe=PromptLine,
  colbacktitle=PromptBlue,
  coltitle=white,
  fonttitle=\normalfont\bfseries\small,
  fontupper=\normalfont\normalsize,
  title={#1},
  boxrule=0.4pt,
  arc=1mm,
  left=3mm, right=3mm,
  top=2mm, bottom=2mm,
  toptitle=1mm, bottomtitle=1mm,
  before upper=\raggedright
}

\newcommand{\ms}[2]{%
  \ensuremath{#1\,{\scriptstyle (#2)}}%
}

\newcommand{\bestms}[2]{%
  \ensuremath{\mathbf{#1}\,{\scriptstyle(#2)}}%
}

\usepackage{xspace}
\newcommand{\ours}{\textit{RubricDuo}\xspace}
\newcommand{\direct}{\textit{Direct-Q}\xspace}
\newcommand{\random}{\textit{Random-FS}\xspace}
\newcommand{\rag}{\textit{RubricRAG}\xspace}
\newcommand{\ret}{\textit{Rubric-Ret}\xspace}
\newcommand{\carg}{\textit{CaRG}\xspace}
\newcommand{\no}{\textit{No-Rubric}\xspace}
\newcommand{\human}{\textit{Human-Rubric}\xspace}

\definecolor{GraphBlue}{HTML}{EEF4FF}
\definecolor{TableRule}{HTML}{B7BDC8}
\definecolor{MutedText}{HTML}{777777}

\usepackage{amsmath,amsfonts,bm}

\def\eqref#1{equation~\ref{#1}}

\def\1{\bm{1}}

\DeclareMathAlphabet{\mathsfit}{\encodingdefault}{\sfdefault}{m}{sl}
\SetMathAlphabet{\mathsfit}{bold}{\encodingdefault}{\sfdefault}{bx}{n}

\title{From Cases to Criteria: Reusing Evaluation Knowledge for Task-Specific Rubric Generation}

\author{%
  Qiyan Deng\textsuperscript{1},
  Long-Kai Huang\textsuperscript{2},
  Kangfei Zhao\textsuperscript{1} \\
  \textsuperscript{1}Beijing Institute of Technology \\
  \textsuperscript{2}Hong Kong Baptist University
}

\begin{document}

\maketitle

\begin{abstract}

Query-specific rubrics make LLM evaluation more fine-grained and transparent, and generating high-quality rubrics requires both identifying the target query’s evaluation scope and specifying concrete checks. Historical rubrics provide reusable evaluation experiences, yet existing approaches reuse either complete query–rubric cases or aggregated criteria and dimensions without explicitly connecting recurring evaluation requirements to their criterion-level evidence and source-query context. This can preserve irrelevant task details or lose the broader evaluation structure.
In this paper, we introduce \ours, which organizes historical query–rubric pairs into a Rubric Knowledge Graph linking evaluation dimensions, shared concepts, local concepts, criteria, and queries, where reusable evaluation requirements across queries are abstracted by shared concepts. 
Given a target query, \ours first combines concepts relevant to the target query with retrieved historical cases to draft a rubric with broad coverage. It then revisits each draft criterion, retrieves concept-linked historical criteria and their source queries, and adapts applicable evidence to refine the criterion.
Experiments on RubricBench and HealthBench demonstrate that \ours achieves the overall best performance among the evaluated automatic methods in both alignment with human-authored rubrics and downstream LLM judging.

\end{abstract}

\section{Introduction}

Large language models (LLMs) are increasingly evaluated by other LLMs acting as judges~\citep{llm-as-a-judge,llm-as-a-judge-2}, especially for complex, open-ended tasks. However, a judge that outputs only a score~\citep{lee-etal-2024-unleashing} or a preference~\citep{ICLR2024_pandalm} reveals little about why one response is better than another.
Rubric-based evaluation makes this reasoning explicit by decomposing the requirements of a query into criteria that can be checked individually~\citep{llm-rubric,ICLR2024_Prometheus,cook2024tickingboxesgeneratedchecklists}. 
Expert-written rubrics now underpin benchmarks such as HealthBench~\citep{arora2025healthbenchevaluatinglargelanguage} and will also serve as reward signals for training LLMs in domains without verifiable answers~\citep{ICLR2026_rubric-as-reward}.
Since writing a rubric for every query requires substantial expert effort, a growing line of work seeks to generate rubrics automatically.

A central challenge in automatic rubric generation is deciding what criteria should check.
On RubricBench~\citep{zhou-etal-2026-rubricbench}, replacing model-generated rubrics with human-written ones raises judge accuracy by about 27 points, and the gap persists as more rubrics are sampled. 
Historical human rubrics are a natural source of the missing knowledge because they record which information experts checked for related queries. 
Recent methods therefore reuse this historical knowledge, retrieving query--rubric pairs similar to the target query as demonstrations~\citep{rubricrag} or by distilling historical criteria into pools of evaluation dimensions or criteria~\citep{mao2026carg,jia-etal-2026-autorubric}. Nevertheless, such reuse fails to reliably improve the quality of the generated rubrics. 
In our experiments (\cref{sec:experiment}), contextualizing the generator with retrieved similar cases fails to improve alignment with human rubrics compared to generating from the query alone in two of five settings.

\begin{figure}[!t]
    \centering
    \includegraphics[width=\linewidth]{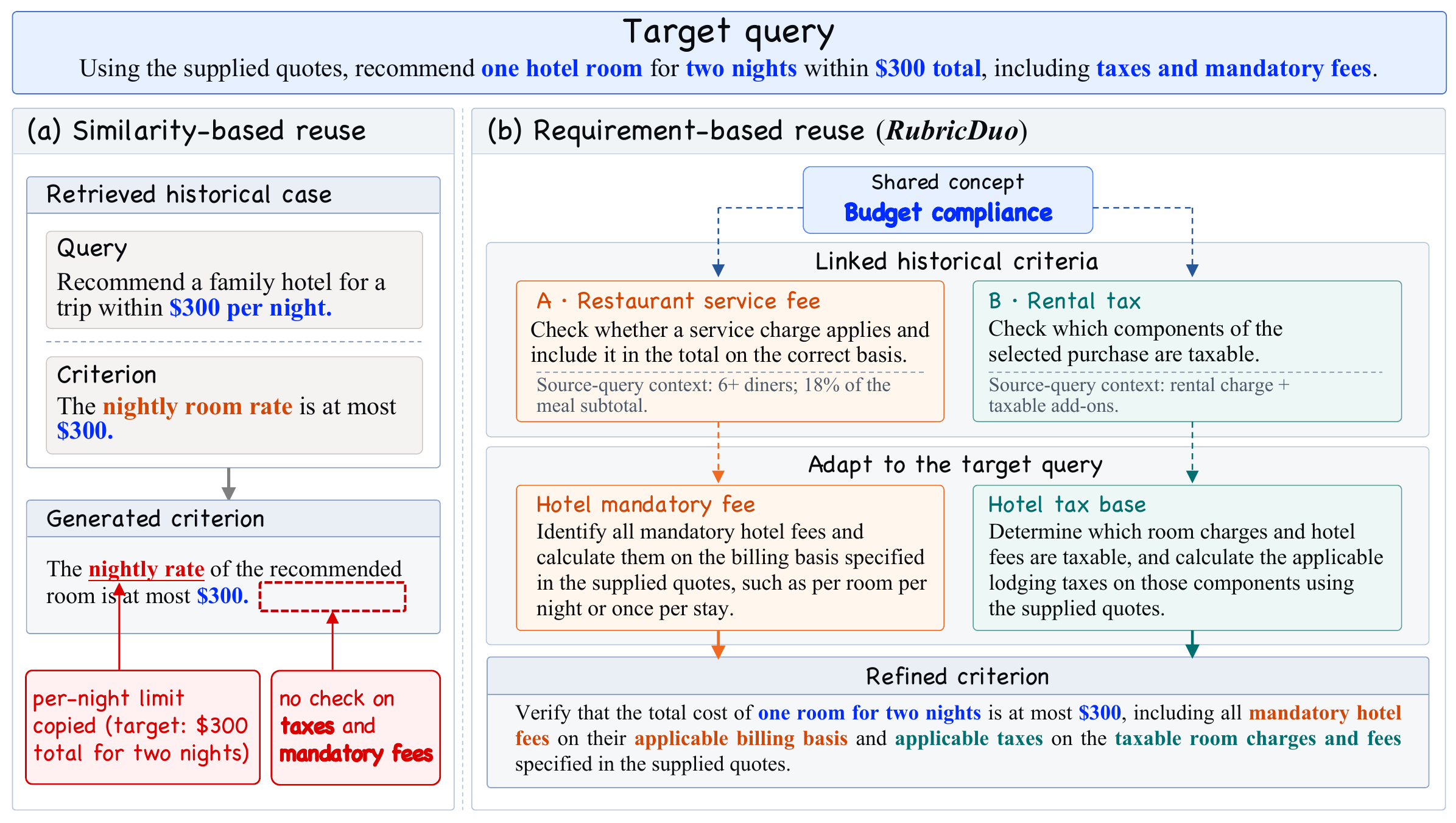}
    \caption{Similarity- and requirement-based rubric reuse. (a) A similar hotel case carries over an incorrect nightly limit and misses taxes and mandatory fees. (b) A shared concept links criteria from different domains to their source query contexts; applicable checks are adapted to verify the \$300 two-night total, including fees and taxes.}
    \label{fig:rubric-reuse}
\end{figure}

We argue that this unreliability stems from the unit at which rubric knowledge is represented and reused. What transfers across queries is usually not a whole rubric or the literal wording of a criterion, but the evaluation requirement that a criterion instantiates at the conceptual level.
Consider the hotel recommendation query in Fig.~\ref{fig:rubric-reuse}(a), which asks for a hotel room for two nights within \$300 in total, including taxes and mandatory fees. Similarity-based reuse favors other hotel queries, such as a request for a family hotel within \$300 per night, whose rubric bounds the nightly rate rather than the total and says nothing about mandatory fees. 
The missing check appears instead in the rubric of a restaurant reservation, which requires the total to include a service charge for large parties, and a car-rental rubric adds a check on which charges are taxable. These queries differ in topic, but their criteria express the same underlying requirement, i.e., the reported total must include every applicable charge. 
Beyond similar case retrieval, effective reuse therefore entails a well-refined abstraction and organization of recurring evaluation requirements across historical cases with various topics. 
First, these evaluation requirements should be abstracted from historical rubrics regarding different evaluation dimensions. 
Second, historical criteria and queries should be indexed with these evaluation requirements, enabling the generator to retrieve knowledge based on the requirements of target queries  while retaining the source context needed to adapt that knowledge.
Existing methods do not provide such a joint knowledge abstraction and organization.
Case retrieval preserves source context but operates at the query level, while pools of dimensions or aggregated criteria capture recurring patterns without explicitly organizing how an abstract requirement connects back to the historical evidence that instantiates it. 

To this end, we propose \ours in this paper, which reuses historical rubrics through their shared evaluation requirements. 
In the offline phase, \ours organizes historical query--rubric pairs into a Rubric Knowledge Graph. The graph abstracts each criterion into the requirements it checks, 
groups recurring requirements into shared evaluation concepts such as `\textit{budget compliance}', and links every concept back to its source criteria and queries. 
In the online phase, \ours generates a rubric in two steps. In the first step, it selects the shared concepts relevant to the target query, retrieves similar historical cases, and drafts a rubric that covers the overall evaluation scope. Similar cases remain useful at this stage because they show how a set of criteria jointly covers a problem, while the selected concepts indicate which requirements the target query raises. In the second step, \ours revisits each draft criterion, retrieves historical criteria linked to the same concept, and adapts only the details that apply to the target query. Because candidates are filtered by concept rather than by query similarity, this evidence can come from tasks unrelated in topic, as shown in Fig.~\ref{fig:rubric-reuse}(b).

We evaluate \ours on RubricBench~\citep{zhou-etal-2026-rubricbench} and HealthBench~\citep{arora2025healthbenchevaluatinglargelanguage}, using both alignment with held-out human rubrics and downstream LLM-judge accuracy. Across five settings covering three rubric generators and cross-benchmark transfer from RubricBench to HealthBench, \ours achieves the highest alignment F1 score and the highest judging accuracy in all five settings. On RubricBench, it improves judging accuracy from 47.08\% with query-only rubrics to 53.34\%, closing 22\% of the gap to human-written rubrics. Controlled comparisons further show that the gain comes from how historical knowledge is organized and reused: replacing graph-guided retrieval with flat dense or hybrid retrieval reduces accuracy, while concept-guided drafting alone already outperforms all baselines.


\section{Related Work}

\paragraph{Rubric Generation from Target Queries.}
Query-conditioned methods generate evaluation criteria directly from the requirements of a target query. 
Given the target query and its associated topic,~\citet{writing-bench} generates instance-specific rubrics. 
TICK~\citep{cook2024tickingboxesgeneratedchecklists} and RocketEval~\citep{wei2025rocketeval} identify individual query requirements and express them as respective rubric items. 
Qworld~~\citep{su2026qworld} decomposes a query into scenarios and evaluation perspectives, then develops fine-grained criteria through recursive tree-based exploration.
These methods improve how generators identify and express evaluation requirements, while our work focuses on how historical evaluation knowledge can be organized and supplied to the generator, with explicit connections between reusable concepts and concrete query-rubric instances.


\paragraph{Rubric Construction from Historical Experience.}
Historical evaluation experience can be reused for new queries through complete cases, shared candidate pools, or criteria  aggregated across queries. 
RubricRAG~\citep{rubricrag} retrieves semantically similar historical query–rubric pairs as few-shot demonstrations to guide rubric generation for a target query. 
CaRG~\citep{mao2026carg} induces a reusable evaluation dimension pool from historical queries and selects pertinent dimensions to guide generation.
Auto-Rubric~\citep{xie2026autorubriclearningimplicitweights} and SibylSense~\citep{xu2026sibylsenseadaptiverubriclearning} derive criteria from comparisons between preferred and dispreferred responses and aggregate these criteria across queries for repurposing on new queries. 
Specifically, Auto-Rubric distills the aggregated criteria into a compact, query-agnostic coreset, while SibylSense organizes the criteria by category. 
Prior studies~\citep{lyu2026clinalign,wang2026infimedorbit} demonstrate different ways to reuse evaluation experience, but only from either the query-side or the criterion-side. 
In contrast, \ours connects the two through shared evaluation concepts as the provenance of cross-query evaluation requirements. This provenance allows complete historical cases to guide rubric coverage while enabling individual historical criteria to provide concrete evidence.

\section{\ours: Methodology}
\label{sec:method}
\begin{figure}[!t]
    \centering
    \includegraphics[width=\linewidth]{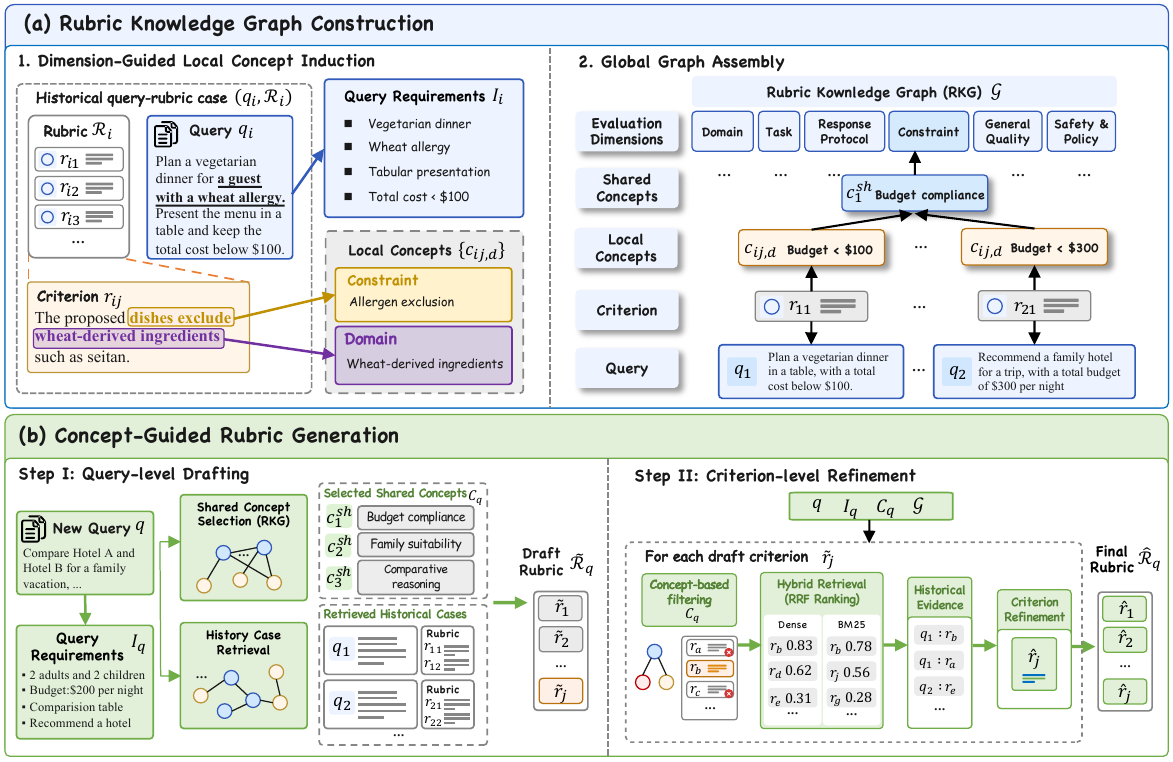}
    \caption{The overview of \ours: (a) The offline phase constructs a Rubric Knowledge Graph guided by six dimension; (b) the online phase generates rubric for a target query via query-level drafting and criterion-level refinement. }
    \label{fig:overview}
\end{figure}
\paragraph{Problem Setup.} Let $\mathcal{H}=\{(q_i,R_i)\}_{i=1}^{N}$ denote a collection of historical queries and their associated rubric,
where the rubric for query $q_i$ contains a set of criteria, i.e., $R_i=\{r_{ij}\}_{j=1}^{m_i}$, for evaluating a response to $q_i$ across $m_i$ facets. 
Throughout this paper, \emph{rubric} refers to the complete set of criteria for a query, while \emph{criterion} refers to an individual item in the rubric. 
Given a new query $q$, our goal is to generate a rubric $\widehat{\mathcal{R}}_q$ that captures its evaluation requirements comprehensively and precisely. 


\paragraph{Overview.} Fig.~\ref{fig:overview} delineates an overview of \ours, which comprises two phases: offline multi-level \emph{Rubric Knowledge Graph} (RKG) construction (\cref{sec:graph_construction}) and online concept-guided rubric generation (\cref{sec:rubric_generation}). 
In the offline phase, \ours constructs an RKG $\mathcal{G} = (\mathcal{V}, \mathcal{E})$ from historical evaluation cases $\mathcal{H}$. 
The graph connects original queries and criteria with derived local concepts, shared concepts, and evaluation dimensions in a level-by-level fashion.  More specifically, to induce concepts, \ours specifies six different general evaluation dimensions from criteria, including Domain, Task, Response Protocol, Constraint, General Quality, and Safety \& Policy.
The constructed multi-level RKG preserves provenance at multiple granularities, linking transferable evaluation dimensions and concepts to concrete historical criteria and queries. 
In the online phase, \ours generates a rubric for a new query $q$ using a two-step, concept-guided strategy.
The first step performs coarse-grained drafting to cover general evaluation aspects of $q$, and then the second step performs fine-grained refinement to make each criterion concrete and consistent. 
Query-level drafting retrieves similar queries and their rubrics and uses them as demonstrations to produce an initial rubric by prompting an LLM generator. 
Subsequently, criterion-level refinement revisits each criterion in the initial rubric. For each criterion, \ours retrieves a set of historical criteria in $\mathcal{G}$ by concept-guided filtering and reranking, and then refers them as detailed evidence to refine the draft criterion, yielding the final rubric.

\subsection{Rubric Knowledge Graph Construction}
\label{sec:graph_construction}

\ours represents the Rubric Knowledge Graph (RKG) as a multi-level heterogeneous graph $\mathcal{G} = (\mathcal{V}, \mathcal{E})$. The graph contains five types of nodes, i.e., $\mathcal{V}
    =
    \mathcal{V}_{Q}
    \cup
    \mathcal{V}_{R}
    \cup
    \mathcal{V}_{C}^{\mathrm{loc}}
    \cup
    \mathcal{V}_{C}^{\mathrm{sh}}
    \cup
    \mathcal{V}_{D}$, where $ \mathcal{V}_{Q}$ and $\mathcal{V}_{R}$ denote the historical queries and individual criteria from $\mathcal{H}$, respectively. $\mathcal{V}_{C}^{\mathrm{loc}}$ and  $\mathcal{V}_{C}^{\mathrm{sh}}$ denote the local and shared concepts induced from $\mathcal{H}$, while $\mathcal{V}_D$ denotes six general evaluation dimensions.
    Intuitively, local concepts $\mathcal{V}_{C}^{\mathrm{loc}}$ capture distinct evaluation requirements delivered by a criterion, while shared concepts $\mathcal{V}_{C}^{\mathrm{sh}}$ represent semantic-level evaluation requirements that recur across multiple criteria and are organized by their corresponding evaluation dimensions $\mathcal{V}_{D}$.

\subsubsection{Dimension-guided Local Concept Extraction}
\label{sec:local_concept_induction}

\ours defines six general evaluation dimensions $\mathcal{V}_{D}$ to guide local concept extraction and shared concept induction, enabling fine-grained reuse of historical experience. These dimensions distinguish six evaluation requirements within a criterion: (1) \emph{Domain} for domain-specific factual or substantive content; (2) \emph{Task} for the objective or operation that the response must perform; (3) \emph{Response Protocol} for requirements concerning format, organization, or presentation; (4) \emph{Constraint} for explicit limits, conditions, or resources; (5) \emph{General Quality} for cross-task qualities such as clarity, coherence, and completeness; (6) \emph{Safety \& Policy} for safety, ethical, legal, or policy-related requirements.

For each historical query $q_i$, we first extract its intent and requirements, i.e., ${I}_i := \operatorname{IntentExtract}(q_i)$. 
Given the source query $q_i$ and its extracted requirements ${I}_i$, we identify the dimensions presented within each criterion $r_{ij} \in \mathcal{R}_i$ as ${D}_{ij} :=
    \operatorname{DimensionIdentify}
    (r_{ij} \mid q_i, {I}_i),
    {D}_{ij}\subseteq \mathcal{V}_{D}$, where $D_{ij}$ is a subset of the six dimensions corresponding to the general aspects evaluated by criterion $r_{ij}$.
The query intent extraction and dimension identification are performed by prompting an LLM with the corresponding contexts. 

For each identified dimension $d \in D_{ij}$, we then extract a local concept $c_{ij,d}$ from $r_{ij}$ that captures a concrete evaluation aspect within $r_{ij}$. 
The extraction is likewise performed through LLM prompting, conditioned on the query $q_i$ and its intent $I_i$, criterion $r_{ij}$, and a dimension $d$, i.e., $c_{ij, d} := \operatorname{ConcepetExtract}
    (r_{ij}\mid q_i, {I}_i, d)$. The local concepts extracted from all queries and their criteria form the local concept entities $\mathcal{V}_{C}^{\mathrm{loc}}$. 

For illustration, consider the criterion in Fig.~\ref{fig:rubric-reuse}: `\textit{The dinner total is at most \$180, including the mandatory 18\% service charge on the pre-tax meal subtotal because the party meets the six-person threshold.}'
This criterion evaluates both the spending constraint and the restaurant-specific fee rule.
Under \emph{Constraint}, it can yield local concepts such as \textit{dinner spending limit}; under \emph{Domain}, it can yield \textit{restaurant service-fee rule}.
The dimensions thus guide the induction of distinct local concepts from the same criterion, while each concept retains its specific evaluation focus and source context.

\subsubsection{Shared Concept Induction \& Graph Assembly}

\paragraph{Shared Concept Induction.}
\ours aggregates semantically relevant local concepts $\mathcal{V}_{C}^{\mathrm{loc}}$ and induces a set of shared concepts $\mathcal{V}_{C}^{\mathrm{sh}}$, aiming to enable the explicit reuse of evaluation requirements within  criteria across queries. 
Specifically, we group all extracted local concepts $\mathcal{V}_{C}^{\mathrm{loc}}$ from $\mathcal{H}$ by their dimension $d$, and prompt an LLM for each group to induce shared concepts for each dimension-specific group.
When a group of local concepts exceeds the effective context window of the LLM, we divide the group into batches and process one batch per prompt. 
We maintain the shared concepts induced across batches and dimensions by merging semantically equivalent concepts and incrementally adding new concepts, yielding the final shared concept set $\mathcal{V}_{C}^{\mathrm{sh}}$.

For example, one criterion requires a dinner plan to cost less than \$100, while another requires a travel itinerary to stay within a \$200 budget. The local concepts induced from these criteria under Constraint both concern staying within a specified budget. We group them under the shared concept `\textit{budget compliance}', while retaining the different tasks and budget limits in their source criteria.

\paragraph{Graph Assembly.} With the six dimensions $\mathcal{V}_D$, the extracted local concepts $\mathcal{V}_{C}^{\mathrm{loc}}$, and the induced shared concepts $\mathcal{V}_{C}^{\mathrm{sh}}$, \ours assembles these entities together with the original queries $ \mathcal{V}_{Q}$ and  rubrics $\mathcal{V}_{R}$ into a multi-level RKG $\mathcal{G} = (\mathcal{V}, \mathcal{E})$, as shown in Fig.~\ref{fig:overview}. These five levels of entities are connected by four types of edges, i.e., $ \mathcal{E} = \mathcal{E}_{q \rightarrow r} \cup \mathcal{E}_{r \rightarrow \text{loc}} \cup \mathcal{E}_{\text{loc} \rightarrow \text{sh}} \cup \mathcal{E}_{\text{sh} \rightarrow d}$. Here, $\mathcal{E}_{q \rightarrow r}$ links each historical query to its original criterion; 
$\mathcal{E}_{r \rightarrow \text{loc}}$ links each criterion to its extracted local concepts; $\mathcal{E}_{\text{loc} \rightarrow \text{sh}}$ links each local concept to its corresponding shared concept; and $\mathcal{E}_{\text{sh} \rightarrow d}$ links each shared concept to its evaluation dimension. 
These connections allow each shared concept to provide access to both individual criteria and their complete historical cases.

\subsection{Concept-Guided Rubric Generation}
\label{sec:rubric_generation}

Given a new query $q$, \ours generates a query-specific rubric $\widehat{\mathcal{R}}_q$ by retrieving and exploiting relevant historical evaluation cases and shared concepts in the RKG $\mathcal{G}$. Generation proceeds in two steps. First, coarse-grained generation drafts an initial rubric $\widetilde{\mathcal{R}}_q$ that covers the general evaluation aspects of $q$ and similar historical queries.
Fine-grained refinement then revises each criterion in $\widetilde{\mathcal{R}}_q$ individually  to produce concrete and consistent criteria as the final rubric $\widehat{\mathcal{R}}_q$. 

\paragraph{Step-I: Query-level Drafting.}

This step first extracts the intent and requirements of the target query $q$, i.e., ${I}_q := \operatorname{IntentExtract}(q)$. Subsequently, an LLM is prompted to select shared concepts in $\mathcal{V}^{\text{sh}}_{C}$ that are relevant to the query intent, such as ${C}_q := \operatorname{ConceptIdentify}(q,{I}_q;\mathcal{G}),{C}_q \subseteq \mathcal{V}_{C}^{\mathrm{sh}}$, which identifies a subset of shared concepts $C_q = \{ c^{\text{sh}}_1, c^{\text{sh}}_2, \cdots \}$ that match the intents of $q$.
Meanwhile, \ours retrieves historical query-rubric cases whose queries are semantically and lexically similar to the target query $q$. Historical queries are ranked using both dense semantic retrieval and BM25 retrieval, and the two rankings are combined using Reciprocal Rank Fusion~\citep{RRF}. 
The top-$K$ retrieved cases, $\mathcal{H}_q = \{ (q_k, \mathcal{R}_k)\}_{k = 1}^{K}$, demonstrate how multiple criteria jointly cover an evaluation problem, while the selected shared concepts $C_q$ indicate which evaluation requirements are relevant to the target query. 
We provide the target query $q$, its extracted intent $I_q$, the selected concepts $C_q$, and the retrieved cases $\mathcal{H}_q$ for the rubric generator to draft the initial rubric:
\begin{align}
    \widetilde{\mathcal{R}}_q
    :=
    \operatorname{Draft}
    (q, {I}_q, {C}_q, \mathcal{H}_q). 
\end{align}
The resulting draft $\widetilde{\mathcal{R}}_q$ is intended to provide broad coverage before criterion-specific evidence is introduced.

\paragraph{Step-II: Criterion-level Refinement.}
For each draft criterion $ \widetilde{r}_j \in \widetilde{\mathcal{R}}_q$, \ours first identifies what the criterion evaluates and maps it to the relevant shared concepts in $C_q$. These concepts are used to filter historical criteria in the graph $\mathcal{G}$, and those aligned with the relevant concepts are retained as a candidate reference pool for criterion $\widetilde{r}_j$. 
Within this candidate pool, hybrid retrieval combines dense and BM25 rankings using Reciprocal Rank Fusion to retrieve criteria with similar terms, conditions, or format requirements to $\widetilde{r}_j$. 
The resulting criteria, along with their source queries, constitute the specific reference evidence $E_j$ for $\widetilde{r}_j$.
The rubric generator evaluates this evidence $E_j$ in the context of the target query $q$ and its intent $I_q$ to refine the draft criterion: 
\begin{align}
    \widehat{r}_j := \operatorname{Refine}(\widetilde{r}_j; q, I_q, E_j) \quad \forall \widetilde{r}_j \in \widetilde{\mathcal{R}}_q. 
\end{align}
The final rubric is obtained by refining every draft criterion and assembling the resulting criteria $\widehat{R}_q$.

\section{Experiment}
\label{sec:experiment}
We give the experimental setup (\cref{sec:experimental-setup}) and evaluate \ours by investigating the following research questions. 
\textbf{RQ-1}: Does \ours generate rubrics that better recover the evaluation requirements from human annotators and improve the downstream LLM judges? (\cref{sec:main-results})
\textbf{RQ-2}: Can an RKG constructed from historical query-rubric pairs support knowledge transfer across benchmarks for rubric generation? (\cref{sec:cross-benchmark-transfer})
\textbf{RQ-3}: Does organizing historical query-rubric observations as a graph enable more effective rubric construction than flat retrieval? (\cref{sec:graph-versus-flat})
\textbf{RQ-4}: Is the two-step generation strategy of \ours effective? (\cref{sec:two-stage-ablation})
\textbf{RQ-5}: Does structured historical evaluation knowledge remain effective across rubric generators of different capabilities? (\cref{sec:generator-capability})

\subsection{Experimental Setup}
\label{sec:experimental-setup}

\textbf{Benchmarks.}
We evaluate rubric generation on RubricBench~\citep{zhou-etal-2026-rubricbench} and HealthBench~\citep{arora2025healthbenchevaluatinglargelanguage}. 
RubricBench contains 1,147 queries with human-authored atomic rubrics, paired responses, and human preference labels, spanning chat, instruction following, STEM, coding, and safety tasks; we use 917 query-rubric pairs as historical observations and reserve 230 examples for held-out evaluation.
HealthBench is curated by OpenAI in collaboration with 262 physicians across 60 countries, pairing realistic healthcare conversations with human-authored rubrics. We use 1,000 query-rubric pairs as historical observations and evaluate on 255 held-out queries using the response pairs constructed by \rag~\citep{rubricrag}.

\textbf{Baselines.}
We compare \ours with five baselines for automatic rubric construction and include two reference conditions.
\no evaluates candidate responses without an explicit rubric, while \human provides the held-out human-authored rubric to the same judge as an oracle reference.
Among the automatic baselines, \direct generates rubrics from the query alone; 
\random conditions generation on randomly sampled historical query-rubric pairs; 
\rag uses semantically similar historical query-rubric pairs as generation exemplars;
\ret uses semantically similar historical criteria as generation exemplars;
and \carg constructs rubrics from a retrieved pool of evaluation dimensions.

\textbf{Weight Assignment.}
RubricBench uses unweighted criteria, whereas HealthBench assigns signed point values.
Because some methods, such as \carg, generate criteria without weights, we use a shared LLM-based weight assigner (Qwen3.7-flash) on HealthBench. The assigner takes the query and generated rubrics as input and assigns a signed point value to each criterion, ensuring a consistent weighting protocol across methods.

\textbf{Metrics.}
Following prior work~\citep{zhou-etal-2026-rubricbench,rubricrag}, we evaluate Rubric Alignment using Rubric Precision, Rubric Recall, and Structure F1 against held-out human-authored rubrics. For downstream utility, we report pairwise accuracy (Acc.) on RubricBench, the proportion of cases where the judge selects the human-preferred response; and good-over-bad accuracy (G$>$B Acc.) on HealthBench, the proportion of pairs where the good response scores strictly higher than the bad response under the generated rubric and its criterion weights.
All methods use the same generator model and fixed LLM judge, history-based methods also share the same historical corpus. 

\textbf{Models and Settings.}
In the main experiments, all methods use Qwen3.7-flash
for rubric generation and weight assigner, and Deepseek-V4.1-flash serves as the downstream judge.
History-based methods share the same historical corpus.
We use Qwen3-enbedding-4b for dense retrieval.
We repeat each automatic method 3 times and report the mean and sample standard deviation.

\subsection{Main Results}
\label{sec:main-results}

\begin{table}[!t]
\caption{Main results on RubricBench and HealthBench.
Values are percentages: mean (sample SD) over three runs.}
\label{tab:main-results}
\centering
\scriptsize
\setlength{\tabcolsep}{3pt}
\renewcommand{\arraystretch}{1.10}
\begin{tabular}{lcccccccc}
\toprule
& \multicolumn{4}{c}{\textbf{RubricBench}}
& \multicolumn{4}{c}{\textbf{HealthBench}} \\
\cmidrule(lr){2-5}
\cmidrule(lr){6-9}
\textbf{Method}
& \multicolumn{3}{c}{\textbf{Alignment}}
& \multicolumn{1}{c}{\textbf{Utility}}
& \multicolumn{3}{c}{\textbf{Alignment}}
& \multicolumn{1}{c}{\textbf{Utility}} \\
\cmidrule(lr){2-4}
\cmidrule(lr){5-5}
\cmidrule(lr){6-8}
\cmidrule(lr){9-9}
& P $\uparrow$ & R $\uparrow$ & F1 $\uparrow$ & Acc. $\uparrow$
& P $\uparrow$ & R $\uparrow$ & F1 $\uparrow$ & G$>$B Acc. $\uparrow$ \\
\midrule
\no
& -- & -- & -- & \ms{46.96}{1.33}
& -- & -- & -- & -- \\
\human
& -- & -- & -- & \ms{75.11}{1.17}
& -- & -- & -- & \ms{56.08}{1.48} \\
\midrule
\direct
& \ms{41.20}{1.30} & \ms{39.97}{1.22} & \ms{39.08}{1.21} & \ms{47.08}{1.70}
& \ms{29.51}{0.81} & \ms{15.39}{0.40} & \ms{18.77}{0.53} & \ms{40.00}{0.42} \\
\random
& \ms{36.72}{0.90} & \ms{40.08}{1.17} & \ms{37.07}{1.02} & \ms{47.08}{0.34}
& \ms{30.07}{0.58} & \ms{21.83}{0.31} & \ms{23.54}{0.36} & \ms{45.46}{3.70} \\
\rag
& \ms{37.87}{0.45} & \ms{40.76}{0.33} & \ms{38.12}{0.35} & \ms{50.65}{2.29}
& \ms{28.37}{0.36} & \ms{\textbf{26.66}}{0.25} & \ms{25.48}{0.26} & \ms{47.63}{1.36} \\
\ret
& \ms{38.53}{0.52} & \ms{39.28}{0.73} & \ms{37.46}{0.55} & \ms{47.15}{1.22}
& \ms{32.85}{0.44} & \ms{18.25}{0.36} & \ms{21.85}{0.45} & \ms{43.01}{0.80} \\
\carg
& \ms{37.71}{1.11} & \ms{36.15}{1.13} & \ms{35.48}{1.07} & \ms{50.88}{0.59}
& \ms{29.69}{1.83} & \ms{15.37}{1.09} & \ms{18.82}{1.29} & \ms{41.28}{1.72} \\
\ours
& \ms{\textbf{43.83}}{1.60} & \ms{\textbf{45.97}}{1.79}
& \ms{\textbf{43.64}}{1.60} & \ms{\textbf{53.34}}{1.09}
& \ms{\textbf{33.21}}{0.62} & \ms{23.49}{0.61} & \ms{\textbf{25.62}}{0.49} & \ms{\textbf{52.03}}{1.41} \\
\bottomrule
\end{tabular}
\end{table}

\noindent \textbf{Rubric Alignment.}
Table~\ref{tab:main-results} shows that \ours achieves the best rubric alignment among all methods on RubricBench, with precision of 43.83\%, recall of 45.97\%, and F1 of 43.64\%.
The higher recall indicates that \ours covers more requirements in the human-authored rubrics, while the higher precision indicates that a larger proportion of its generated criteria match the reference criteria.
Moreover, retrieving historical examples alone does not consistently improve alignment: both \random and \rag obtain lower F1 than \direct on RubricBench.
This suggests that access to historical rubrics is insufficient without effective selection and adaptation to the current query.
On HealthBench, RubricDuo achieves the highest precision (33.21\%) and Structure F1 (25.62\%) among all methods,

\noindent \textbf{Downstream Utility.}
On RubricBench, \ours achieves the highest accuracy among all methods at 53.34\%, improving upon \rag by 2.69 percentage points and \no by 6.38 percentage points.
On HealthBench, \ours also achieves the highest G$>$B accuracy among all methods at 52.03\%.
However, gaps to \human remain: 21.77 percentage points on RubricBench and 4.05 percentage points on HealthBench. 

We also observe that \carg achieves lower F1 than \direct but higher downstream accuracy on RubricBench.
This indicates that stronger agreement with human-authored rubrics does not necessarily translate into better downstream performance, as criteria useful for distinguishing response quality may extend beyond those included in the reference rubric.
The improvements of \ours in both F1 and accuracy across the two benchmarks suggest that its generated rubrics support both the recovery of human-authored evaluation requirements and effective downstream response evaluation.

\subsection{Cross-Benchmark Transfer}
\label{sec:cross-benchmark-transfer}










\begin{table}[t]
\caption{Cross-benchmark transfer from RubricBench History to HealthBench.}
\label{tab:cross-benchmark-transfer}
\centering
\footnotesize
\setlength{\tabcolsep}{3.2pt}
\renewcommand{\arraystretch}{1.10}

\begin{tabular}{lcccc}
\toprule
\textbf{Method}
& \textbf{P} $\uparrow$
& \textbf{R} $\uparrow$
& \textbf{F1} $\uparrow$
& \textbf{G$>$B Acc.} $\uparrow$ \\
\midrule

\direct
& \ms{30.08}{0.49}
& \ms{15.47}{0.27}
& \ms{18.98}{0.33}
& \ms{41.40}{0.44} \\

\random
& \ms{28.40}{0.66}
& \ms{16.26}{0.77}
& \ms{19.19}{0.60}
& \ms{42.73}{1.27} \\

\rag
& \ms{26.24}{0.51}
& \ms{15.12}{0.22}
& \ms{17.77}{0.34}
& \ms{44.84}{1.13} \\

\ret
& \ms{27.58}{0.71}
& \ms{15.13}{0.34}
& \ms{17.94}{0.39}
& \ms{40.94}{0.91} \\

\carg
& \ms{29.06}{0.85}
& \ms{13.36}{0.46}
& \ms{16.92}{0.57}
& \ms{42.01}{3.02} \\

\ours
& \ms{\textbf{30.92}}{0.61}
& \ms{\textbf{16.94}}{0.83}
& \ms{\textbf{20.38}}{0.62}
& \ms{\textbf{47.90}}{1.31} \\

\bottomrule
\end{tabular}
\end{table}

Table~\ref{tab:cross-benchmark-transfer} evaluates whether historical evaluation knowledge organized in a rubric graph can transfer across benchmarks and support rubric construction under
a shift in task distribution.
We construct the graph from RubricBench history and apply it to HealthBench queries without using HealthBench historical cases.
All methods show limited alignment with human-authored rubrics, achieving Structural F1 scores of approximately 17\%. This is because experts across benchmarks may follow different professional standards, RubricBench covers a broad range of general-purpose tasks, whereas HealthBench focuses on specialized healthcare scenarios that require more clinically specific evaluation criteria. Consequently, directly using RubricBench examples to guide rubric generation for HealthBench queries provides limited benefit and may even reduce downstream accuracy.
%
\ours achieves the highest Structure F1 and accuracy, suggesting that structured knowledge reuse remains valuable for evaluating responses across benchmarks, even when alignment with target expert rubrics is limited.

\begin{table}[t]
\caption{
Comparison of \textit{Flat-Dense}, \textit{Flat-Hybrid}, and \textit{Graph-Guided} retrieval
within the same draft-then-refine framework on RubricBench.
}
\label{tab:graph-versus-flat}
\vspace{3pt}
\centering
\footnotesize
\setlength{\tabcolsep}{5pt}
\renewcommand{\arraystretch}{1.10}

\begin{tabular}{lcccc}
\toprule
Retrieval
& P $\uparrow$
& R $\uparrow$
& F1 $\uparrow$
& Acc. $\uparrow$ \\
\midrule

\textit{Flat-Dense}
& \ms{42.76}{1.36}
& \ms{44.40}{0.54}
& \ms{42.38}{0.92}
& \ms{52.21}{0.60} \\

\textit{Flat-Hybrid}
& \ms{43.74}{0.83}
& \ms{45.71}{1.52}
& \ms{43.51}{1.06}
& \ms{53.00}{1.66} \\

\textit{Graph-Guided}
& \bestms{43.83}{1.60}
& \bestms{45.97}{1.79}
& \bestms{43.64}{1.60}
& \bestms{53.34}{1.09} \\

\bottomrule
\end{tabular}

\end{table}

\subsection{Graph Organization versus Flat Retrieval}
\label{sec:graph-versus-flat}
We examine whether graph-guided retrieval improves rubric refinement over flat dense retrieval on RubricBench.
All variants follow the same draft-then-refine framework.
(1) \textit{Flat-Dense} first retrieves complete historical query-rubric cases based on query-level dense similarity and uses them to generate an initial draft. For each draft criterion, it then retrieves semantically similar historical criteria to support refinement.
(2) \textit{Flat-Hybrid} follows the same two-stage pipeline as
\textit{Flat-Dense}, but combines dense and BM25 retrieval using
reciprocal rank fusion (RRF) to integrate semantic and lexical
matching signals.
(3) \textit{Graph-Guided} is our method, \ours. It follows the same two-stage framework and uses graph-guided retrieval to provide structured historical evidence for rubric refinement.
As shown in Table~\ref{tab:graph-versus-flat}, \textit{Graph-Guided} method achieves an F1 of 43.64\% and an accuracy of 53.34\%, exceeding \textit{Flat-Dense} and \textit{Flat-Hybrid}.
This is because the rubric knowledge graph connects historical cases through shared concepts and evaluation dimensions, enabling the refiner to combine complementary evidence when adapting the rubric to the current query.
\textit{Flat-Hybrid} achieves stronger performance than \textit{Flat-Dense}. This improvement may arise because BM25 complements semantic retrieval with exact lexical matching, allowing the model to retrieve criteria that explicitly specify answer formats and task requirements even when they are not semantically close to the query and may therefore be overlooked by dense retrieval alone.

\subsection{Two-step Draft–refine Ablation}
\label{sec:two-stage-ablation}
\begin{table}[t]
\caption{
Two-step draft-refine ablation on RubricBench.}
\label{tab:two-stage-ablation}
\centering
\footnotesize
\setlength{\tabcolsep}{3.2pt} \renewcommand{\arraystretch}{1.10}
\begin{tabular}{lcccc}
\toprule
Step
& Precision
& Recall
& F1
& Accuracy \\
\midrule

\ours{} (Draft)
& \ms{42.90}{0.54}
& \ms{44.62}{1.28}
& \ms{42.60}{0.84}
& \ms{50.15}{0.26} \\

\ours{} (Refined)
& \ms{43.83}{1.60}
& \ms{45.97}{1.79}
& \ms{43.64}{1.60}
& \ms{53.34}{1.09} \\

\midrule
$\Delta$ (Refined $-$ Draft)
& \textbf{+0.93}
& \textbf{+1.35}
& \textbf{+1.04}
& \textbf{+3.19} \\

\bottomrule
\end{tabular}

\end{table}
To examine the effectiveness of the two-step design, we separately evaluate the draft rubric and the final rubric after criterion-level refinement.
As shown in Table~\ref{tab:two-stage-ablation}, refinement improves all metrics, with gains of 0.93 points in precision, 1.35 in recall, 1.04 in F1, and 3.19 in accuracy.
The strong draft performance suggests that the first step already captures much of the target query's overall evaluation scope, while the consistent gains after refinement show that fine-grained historical criteria provide complementary value by improving the specificity and coverage of individual criteria.

\subsection{Effect of Rubric Generator Capability}
\label{sec:generator-capability}

\begin{table*}[t]
\caption{Effect of rubric-generator capability on RubricBench.
Bold marks the best reported result per row.}
\label{tab:generator-capability}
\centering
\scriptsize
\setlength{\tabcolsep}{3.2pt}
\renewcommand{\arraystretch}{1.10}

\begin{tabular*}{\textwidth}{
    @{\extracolsep{\fill}}
    ll
    !{\color{TableRule}\vrule width 0.45pt}
    ccccc
    >{\columncolor{GraphBlue}}c
    @{}
}
\toprule
\multicolumn{1}{c}{\textbf{Generator}}
& \multicolumn{1}{c}{\textbf{Metric}}
& \textbf{\direct}
& \textbf{\random}
& \textbf{\rag}
& \textbf{\ret}
& \textbf{\carg}
& \textbf{\ours} \\
\midrule

\multirow{4}{*}{Qwen3.5-flash}
& Struct. P $\uparrow$
& \ms{36.05}{1.19}
& \ms{35.65}{0.83}
& \ms{37.97}{0.52}
& \ms{37.28}{0.32}
& \ms{35.24}{0.56}
& \ms{\textbf{40.52}}{0.46} \\

& Struct. R $\uparrow$
& \ms{34.35}{0.87}
& \ms{37.29}{0.48}
& \ms{41.12}{0.82}
& \ms{36.91}{0.70}
& \ms{33.34}{0.09}
& \ms{\textbf{42.27}}{1.32} \\

& Struct. F1 $\uparrow$
& \ms{34.04}{0.92}
& \ms{35.26}{0.58}
& \ms{38.32}{0.64}
& \ms{35.72}{0.53}
& \ms{32.99}{0.24}
& \ms{\textbf{40.19}}{0.86} \\

& Acc. $\uparrow$
& \ms{45.53}{1.55}
& \ms{48.62}{0.34}
& \ms{50.88}{2.37}
& \ms{47.68}{0.25}
& \ms{51.75}{1.09}
& \ms{\textbf{55.03}}{0.91} \\
\midrule

\multirow{4}{*}{Qwen3.7-flash}
& Struct. P $\uparrow$
& \ms{41.20}{1.30}
& \ms{36.72}{0.90}
& \ms{37.87}{0.45}
& \ms{38.53}{0.52}
& \ms{37.71}{1.11}
& \ms{\textbf{43.83}}{1.60} \\

& Struct. R $\uparrow$
& \ms{39.97}{1.22}
& \ms{40.08}{1.17}
& \ms{40.76}{0.33}
& \ms{39.28}{0.73}
& \ms{36.15}{1.13}
& \ms{\textbf{45.97}}{1.79} \\

& Struct. F1 $\uparrow$
& \ms{39.08}{1.21}
& \ms{37.07}{1.02}
& \ms{38.12}{0.35}
& \ms{37.46}{0.55}
& \ms{35.48}{1.07}
& \ms{\textbf{43.64}}{1.60} \\

& Acc. $\uparrow$
& \ms{47.08}{1.70}
& \ms{47.08}{0.34}
& \ms{50.65}{2.29}
& \ms{47.15}{1.22}
& \ms{50.88}{0.59}
& \ms{\textbf{53.34}}{1.09} \\
\midrule

\multirow{4}{*}{Qwen3.8-flash}
& Struct. P $\uparrow$
& \ms{28.65}{0.30}
& \ms{30.59}{0.64}
& \ms{29.59}{0.25}
& \ms{28.62}{0.13}
& \ms{30.81}{0.63}
& \ms{\textbf{37.83}}{0.99}\\

& Struct. R $\uparrow$
& \ms{53.27}{0.42}
& \ms{55.20}{0.82}
& \ms{\textbf{57.32}}{0.67}
& \ms{54.03}{0.66}
& \ms{54.04}{0.50}
& \ms{56.73}{1.01}\\

& Struct. F1 $\uparrow$
& \ms{35.80}{0.48}
& \ms{37.99}{0.67}
& \ms{37.90}{0.05}
& \ms{36.14}{0.24}
& \ms{38.01}{0.56}
& \ms{\textbf{41.85}}{1.05}\\

& Acc. $\uparrow$
& \ms{54.88}{1.35}
& \ms{55.67}{2.04}
& \ms{55.47}{1.21}
& \ms{54.99}{0.46}
& \ms{55.69}{1.47}
& \ms{\textbf{55.85}}{0.95}\\
\bottomrule
\end{tabular*}
\end{table*}

We investigate whether structured historical evaluation knowledge helps less capable models generate higher-quality rubrics. We vary the rubric generator across Qwen3.5-flash, Qwen3.7-flash, and Qwen3.8-flash, while keeping all other settings identical to those of the main RubricBench experiment. As shown in Table~\ref{tab:generator-capability}, \ours achieves the highest Structure F1 with each generator. With Qwen3.5-flash, it also matches or exceeds the Structure F1 of every baseline using Qwen3.8-flash, while attaining competitive Accuracy. These results suggest that structured historical knowledge helps less capable generators identify evaluation requirements they may miss from the query alone.

\section{Conclusion}
Task-specific rubric generation requires both coherent evaluation coverage and concrete checks tailored to the target query. We presented \ours, which organizes historical query–rubric pairs into a Rubric Knowledge Graph that links evaluation dimensions and shared concepts to their source criteria and queries. 
To generate rubrics for a target query, \ours first combines relevant concepts with complete historical cases to draft a rubric, then retrieves concept-linked criterion-level evidence to refine each draft criterion.
Across RubricBench and HealthBench, \ours achieves improvements in alignment with expert-written rubrics and downstream judging against prior automatic baselines. These findings highlight a broader principle for reusing evaluation knowledge: cases help establish what an evaluation should cover, while criteria help specify what it should check. 

\clearpage

\bibliography{iclr2027_conference}
\bibliographystyle{iclr2027_conference}



\end{document}